\documentclass[11pt,a4paper]{article}
\usepackage[hyperref]{naaclhlt2019}
\usepackage{times}
\usepackage{latexsym}

\usepackage{breakurl}
\aclfinalcopy 

\usepackage{relsize}
\usepackage{tabularx}
\usepackage{graphicx}
\usepackage{float}
\usepackage{xspace}
\usepackage{footmisc}
\usepackage{lscape}
\usepackage{makecell}

\title{Multi-Context Term Embeddings: \\the Use Case of Corpus-based Term Set Expansion}

\author{Jonathan Mamou,$^1$ Oren Pereg,$^1$ Moshe Wasserblat,$^1$ Ido Dagan$^2$ \\
  $^1$Intel AI Lab, Israel \\
  $^2$Department of Computer Science, Bar-Ilan University,  Ramat Gan, Israel \\
  $^1${\tt \{jonathan.mamou,oren.pereg,moshe.wasserblat\}@intel.com} \\
  $^2${\tt dagan@cs.biu.ac.il}\\
}

\date{}

\begin{document}

\maketitle
\thispagestyle{empty}
\pagestyle{empty}

\begin{abstract}

In this paper, we present a novel algorithm that combines multi-context term embeddings using a neural classifier and we test this approach on the use case of corpus-based term set expansion. 
In addition, we present a novel and unique dataset for intrinsic evaluation of corpus-based term set expansion algorithms.
We show that, over this dataset, our algorithm provides up to 5 mean average precision points over the best baseline.

\end{abstract}

\section{Introduction}
Term set expansion is the task of expanding a given seed set of terms into a more complete set of terms that belong to the same semantic class. For example, given a seed of personal assistant application terms like `Siri' and `Cortana', the expanded set is expected to include additional terms such as `Amazon Echo' and `Google Now'.

Most prior work on corpus-based term set expansion is based on distributional similarity, where early work is primarily based on using sparse vectors while recent work is based on word embeddings. 
The prototypical term set expansion methods utilize corpus-based semantic similarity between seed terms and candidate expansion terms. To the best of our knowledge, each of the prior methods used a single context type for embedding generation, and there are no reported comparisons of the effectiveness of embedding different context types. 
Moreover, the lack of a publicly available dataset hinders the replicability of previous work and method comparison.

In this paper, we investigate the research question of whether embeddings of different context types can complement each other and enhance the performance of computational semantics tasks like term set expansion. 
To address this question, we propose an approach that combines term embeddings over multiple contexts for capturing different aspects of semantic similarity. The algorithm uses 5 different context types, 3 of which were previously proposed for term set expansion and additional two context types that were borrowed from the general distributional similarity literature. We show that combining the different context types yields improved results on term set expansion.
In addition to the algorithm, we developed a dataset for intrinsic evaluation of corpus-based set expansion algorithms, which we propose as a basis for future comparisons. 

Code, demonstration system, dataset and term embeddings pre-trained models are distributed as part of NLP Architect by Intel AI Lab.~\footnote{\url{http://nlp_architect.nervanasys.com/term_set_expansion.html}}



\section{Related Work}
\label{sec:rel:work}
Several works have addressed the term set expansion problem. We focus on corpus-based approaches based on the distributional similarity hypothesis~\cite{harris1954distributional}. 
State-of-the-art techniques return the $k$ nearest neighbors around the seed terms as the expanded set, where terms are represented by their co-occurrence or embedding vectors in a training corpus according to different context types, such as linear window context~\cite{pantel2009web,shi2010corpus,rong2016egoset,zaheer2017deep,gyllensten2018distributional,Zhao:2018:ESE:3234944.3234966}, explicit lists~\cite{roark1998noun,sarmento2007more,he2011seisa}, coordinational patterns~\cite{sarmento2007more} and unary patterns~\cite{rong2016egoset,shen2017setexpan}. 
In this work, we generalize coordinational patterns, look at additional context types and combine multiple context-type embeddings.



We did not find any suitable publicly available dataset to train and evaluate our set expansion algorithm. The INEX Entity Ranking track~\cite{demartini2009overview} released a dataset for the list completion task. However, it addresses a somewhat different task: in addition to seed terms, an explicit description of the semantic class is supplied as input to the algorithm and is used to define the ground truth expanded set. Some works like~\cite{pantel2009web} provide an evaluation dataset that does not include any training corpus, which is required for comparing corpus-based approaches. \citet{sarmento2007more} use Wikipedia as training corpus, but exploit meta-information like hyperlinks to identify terms; in our work, we opted for a dataset that matches real-life scenarios where terms have to be automatically identified.   

Systems based on our approach are described by~\cite{mamou2018term, mamou2018term2}.

\section{Term Representation}\label{sec:term:rep}

Our approach is based on representing any term in a (unlabeled) training corpus by its word embeddings in order to estimate the similarity between seed terms and candidate expansion terms.
Different techniques for term extraction are described in detail by \citet{moreno2016text}. We follow \citet{kageura1996methods} who approximate terms by noun phrases (NPs),\footnote{Our algorithm can be used for terms with other part-of-speech or with other term extraction methods.} extracting them using an NP chunker. We use {\it term} to refer to such extracted NP chunk and {\it unit} to refer to either a term or a word. 

As preprocessing, term variations, such as aliases, acronyms and synonyms, which refer to the same entity, are grouped together.\footnote{For that, we use a heuristic algorithm based on text normalization, abbreviation web resources, edit distance and word2vec similarity. For example, {\it New York, New-York, NY, NYC} and {\it New York City} are grouped. 
} 
Next, we use term groups as input elements for embedding training (the remaining corpus words are left intact); this enables obtaining more contextual information compared to using individual terms, thus enhancing embedding model robustness. In the remainder of this paper, by language abuse, {\it term} will be used instead of term group.

While word2vec originally uses a linear window context around the {\it focus word}, the literature describes other possible context types. 
For each {\it focus unit}, we extract {\it context units} of different types, as follows (see a typical example for each type in Table~\ref{table:context:example}\footnote{We preferred showing in the example the strength of each context type with a good example, rather than providing a common example sentence across all the context types.}).

\begin{table*}
\begin{tabularx}{\textwidth}{|l|X|X|}
    \hline
    \bf{Cont. type} &  \bf{Example sentence}  & \bf{Context units} \\ \hline
    Lin $win=5$ & {\it {\bf Siri} uses voice queries and a natural language user interface. } & {\it uses, voice queries, natural language user interface}\\ \hline
    List & {\it Experience in {\bf image processing}, signal processing, computer vision.} & {\it signal processing, computer vision} \\ \hline
    Dependency & {\it Turing {\bf studied} as an undergraduate ... at King's College, Cambridge.} & {\it (Turing/nsubj), (undergraduate\-/\-prep\_as), (King's College/prep\_at)} \\ \hline
    SP & {\it {\bf Apple} and Orange juice drink ... }  & {\it Orange} \\ \hline
    UP & {\it In the U.S. state of {\bf Alaska} ...}  & {\it U.S. state of \_\_} \\ \hline
\end{tabularx}
\caption{Examples of extracted context units per context type. Focus units appear in bold. 
}
\label{table:context:example}
\end{table*}

\subsection{Linear Context (Lin)} 
This context is defined by neighboring context units within a fixed length window of context units, denoted by $win$, around the focus unit. 
{\tt word2vec}~\cite{mikolov2013distributed}, {\tt GloVe}~\cite{pennington2014glove} and {\tt fastText}~\cite{joulin2016fasttext} are state-of-the-art implementations.

\subsection{Explicit Lists} Context units consist of terms co-occurring with the focus term in textual lists such as comma separated lists and bullet lists~\cite{roark1998noun,sarmento2007more}. 

\subsection{Syntactic Dependency Context (Dep)} This context is defined by the syntactic dependency relations in which the focus unit participates~\cite{levy2014dependency,macavaney2018deeper}. The context unit is concatenated with the type and the direction of the dependency relation.~\footnote{Given a focus unit $t$ with modifiers $m_i$ and a head $h$, the context of $t$ consists of the pairs $(m_i/l_i)$, where $l_i$ is the type of the dependency relation between the head $h$ and the modifier $m_i$; the context stores also $(h/l_i^{-1})$ where $l_i^{-1}$ marks the inverse-relation between $t$ and $h$.} 
This context type has not yet been used for set expansion. However, \citet{levy2014dependency} showed that it yields more functional similarities of a co-hyponym nature than linear context and thus may be relevant to set expansion.

\subsection{Symmetric Patterns (SP)} Context units consist of terms co-occurring with the focus term in symmetric patterns~\cite{schwartz2015symmetric}. We
follow~\citet{davidov2006efficient} for automatic extraction of SPs from the textual corpus.\footnote{SPs are automatically extracted using the {\tt dr06} library available at \url{https://homes.cs.washington.edu/~roysch/software/dr06/dr06.html}.}
For example, the symmetric pattern `X rather than Y' captures certain semantic relatedness between the terms X and Y. This context type generalizes coordinational  patterns (`X and Y', `X or Y'), which have been used for set expansion.

\subsection{Unary Patterns (UP)} This context is defined by the unary patterns in which the focus term occurs. Context units consist of $n$-grams of terms and other words, where the focus term occurs; `\_\_' denotes the placeholder of the focus term in Table~\ref{table:context:example}. Following~\citet{rong2016egoset}, we extract six $n$-grams per focus term.\footnote{Given a sentence fragment $c_{-3} \; c_{-2} \; c_{-1} \; t \; c_1 \; c_2 \; c_3$ where $t$ is the focus term and $c_i$ are the context units, the following $n$-grams are extracted: 
$(c_{-3} \; c_{-2} \; c_{-1} \; t \; c_1)$, 
$(c_{-2} \; c_{-1} \; t \; c_1 \; c_2)$,
$(c_{-2} \; c_{-1} \; t \; c_1)$,
$(c_{-1} \; t \; c_1 \; c_2 \; c_3)$, 
$(c_{-1} \; t \; c_1 \; c_2)$,
$(c_{-1} \; t \; c_1)$.} 

\paragraph{} 
We show in Section~\ref{sec:exp} that different context types complement each other by capturing different types of semantic relations. 
As explained in Section~\ref{sec:rel:work}, to the best of our knowledge, several of these context types have been used for set expansion, except for syntactic dependency context and symmetric patterns.
We train a separate term embedding model for each of the 5 context types and thus, for each term, we obtain 5 different vector representations.
When training for a certain context type, for each focus unit in the corpus, corresponding {\tt <focus unit, context unit>} pairs are extracted from the corpus and are then fed to the {\tt word2vecf} toolkit that can train embeddings on arbitrary contexts, except for linear context for which we use the {\tt word2vec} toolkit. Only terms representations are stored in the embedding models while other word representations are pruned.

\section{Multi-Context Seed-Candidate Similarity}

For a given context type embedding and a seed term list, we compute two similarity scores between the seed terms and each candidate term, based on cosine similarity.~\footnote{\citet{sarmento2007more} and \citet{pantel2009web} use first-order semantic similarities for explicit list and coordinational pattern context types, respectively. However, \citet{schwartz2015symmetric} showed that for the symmetric patterns context type, word embeddings similarity (second-order) performs generally better. We opted for term embeddings similarity (second-order) for all the context types.}
First, we apply the {\it centroid} scoring method ({\it cent}), commonly
used for set expansion~\cite{pantel2009web}. The centroid of the seed is represented by the average of the term embedding vectors of the seed terms. Candidate terms become the $k$ terms\footnote{\label{notek}Optimal values for $k$ and $k'$ are tuned on the training term list. Other terms are assigned a similarity score of 0 for normalization and combination purpose.}  that are the most similar, by cosine similarity, to the centroid of the seed. 
Second, the {\it CombSUM} scoring method ({\it csum}) is commonly
used in Information Retrieval~\cite{Fox94combinationof}. We first produce a candidate term set for each individual seed term: candidate terms become the $k'$ terms\footref{notek} that are the most similar, according to the term embedding cosine similarity, to the seed term.
The CombSUM method scores the similarity of a candidate term to the seed terms by averaging over all the seed terms the normalized pairwise cosine similarities\footnote{For any seed term, cosine similarities are normalized among the candidate terms in order to combine cosine similarity values estimated on different seed terms for the same candidate term, as suggested by \citet{wu2006evaluating}.} between the candidate term and the seed term.
To combine multi-context embeddings, we follow the general idea of~\citet{berant2012learning} who train an SVM to combine different similarity score features to learn textual entailment relations. Similarly, we train a Multilayer Perceptron (MLP) binary classifier that predicts whether a candidate term should be part of the expanded set based on 10 similarity scores (considered as input features), using the above 2 different scoring methods for each of the 5 context types. 
Note that our MLP classifier polynomially combines different semantic similarity estimations and performs better than their linear combination.
We also tried to concatenate the multi-context term embeddings in order to obtain a single vector representing all the context types. We trained an MLP classifier with concatenated vectors of candidate and seed terms as input features, but it performed worst (see Section~\ref{sec:exp}).

\section{Dataset}
\label{sec:dataset}

Given the lack of suitable standard dataset for training and testing term set expansion models, we used Wikipedia to develop a standard dataset.
Our motivation for using Wikipedia is two-fold. First, Wikipedia contains human-generated lists of terms (`List of' pages) that cover many domains; these lists can be used for supervised training (MLP training in our approach) and for evaluating set expansion algorithms. Second, it contains
textual data that can be used for unsupervised training of corpus-based approaches (multi-context term embedding training in our approach). 
We thus extracted from an English Wikipedia dump a set of term lists and a textual corpus for term embedding training.

\subsection{Term Lists}  
A Wikipedia `List of' page contains terms belonging to a specific class, where a term is defined to be the title of a Wikipedia article.
We selected term lists among `List of' pages containing between fifty and eight hundred terms in order to cover both specific and more common classes (e.g., list of {\it chemical elements} vs. list of {\it countries}). Moreover, we selected term lists that define purely a semantic class, with no additional constraints (e.g., skipping list of {\it biblical names starting with N'}). Since there can be some problems with some Wikipedia `List of' pages, 28 term lists have been validated manually and are used as ground truth in the evaluation. Here are some few examples of term lists: Australian cities, chemical elements, countries, diplomatic missions of the United Kingdom, English cities in the United Kingdom, English-language poets, Formula One drivers, French artists, Greek mythological figures, islands of Greece, male tennis players, Mexican singers, oil exploration and production companies.

Terms having a frequency lower than 10 in the training corpus are pruned from the lists since their embeddings cannot be learned properly; note that these terms are generally less interesting in most of real case applications.
Term variations are grouped according to Wikipedia redirect information. 

On average, a term list contains 328 terms, of which 3\% are not recognized by the noun phrase chunker; the average frequency of the terms in the corpus is 2475.

The set of term lists is split into train, development (dev) and test sets with respectively 5, 5 and 18 lists for MLP training, hyperparameters tuning and evaluation.
Each term list is randomly split into {\it seed} and {\it expanded} term sets, where we are interested in getting enough samples of seed and expanded term sets. 
Thus, given a term list, we randomly generate 15 seed sets (5 seed sets for each seed size of 2, 5 and 10 terms) where seed terms are sampled among the top 30 most frequent terms within the list. For the train set, the non-seed terms (expanded term set) provide the positive samples; we randomly select candidate terms that occur in the corpus but not in the list as negative samples; positive and negative classes are balanced.

\subsection{Textual Corpus} The corpus contains all the textual parts of Wikipedia articles except `List of' pages.~\footnote{Note that the corpus does not contain any Wikipedia meta information.} 
It is used for training the multi-context embedding models. 3\% of the terms appearing in the term
lists are not recognized by our NP chunker in the corpus.
It contains 2.2 billion words, and 12 million unique terms are automatically extracted. 

\subsection{Public Release} 
We use {\tt enwiki-20171201} English Wikipedia dump~\footnote{\url{enwiki-20171201-pages-articles-multistream.xml.bz2}} to develop the dataset.
Full dataset will be released upon publication and it will include train, dev and test sets including the split into seed and expanded terms, and negative samples for the train set; the textual corpus along with NP chunks and grouped term variations; term embedding model for each context type.

\section{Implementation Details}

Code is distributed under the Apache license as part of NLP Architect by Intel AI Lab~\footnote{\url{http://nlp_architect.nervanasys.com/}}, an open-source Python library for exploring  state-of-the-art deep learning topologies and techniques for natural language processing and natural language understanding.

We used the following tools for the implementation and for the development of the dataset:  {\tt spaCy}~\footnote{\url{https://spacy.io}} for tokenization, noun phrase chunking and dependency parsing; {\tt textacy}~\footnote{\url{https://github.com/chartbeat-labs/textacy}} for text normalization; {\tt word2vec}~\footnote{\url{https://code.google.com/archive/p/word2vec}} and {\tt fastText}~\footnote{\url{https://github.com/facebookresearch/fastText}} to model term embeddings of linear context type; {\tt word2vecf}~\footnote{\url{https://bitbucket.org/yoavgo/word2vecf}} to model term embeddings of other context types; {\tt WikiExtractor}~\footnote{\url{https://github.com/attardi/wikiextractor}} to extract textual part of Wikipedia dump; {\tt Keras}~\footnote{\url{https://github.com/keras-team/keras}} to implement the MLP classifier. 

Similarity scores are softmax-normalized over all the candidate terms per context type and per scoring method, in order to combine them with the MLP classifier. Our MLP network consists of one hidden layer. The input and hidden layers have respectively ten and four neurons.

\section{Experiments}\label{sec:exp}

Following previous work~\cite{sarmento2007more}, we report the Mean Average Precision at several top $n$ values (MAP@$n$) to evaluate ranked candidate lists returned by the algorithm. When computing MAP, a candidate term is considered as matching a gold term if they both appear in the same term variations group. 
We first compare the different context types; then, we report results on their combination.

\subsection{Context Type Analysis}
We provide a comparison of the different context types in Table~\ref{tab:context}. 
These context types are baselines and we compare them to the linear context that is more standard. Note that the dependency context type is affected by the performance of the dependency parser.\footnote{We used {\tt spaCy} for dependency parsing; it achieves 92.6\% accuracy on the OntoNotes 5 corpus~\cite{choi2015depends}.}
Linear context with centroid scoring yields consistently best performance of at least 19 MAP@10 points and is consistently more stable looking at standard deviation. However, other context types achieve better performance than linear context type for 55\% of the term lists, suggesting that the different context types complement each other by capturing better different types of semantic relations and that their combination may improve the quality of the expanded set.

In addition, performance consistently increases with the number of seed terms e.g., MAP@10, MAP@20 and MAP@50 of the linear context are respectively .66, .58 and .51 with 2 seed terms.

\begin{table}
    \centering

    \begin{tabular}{l|l|c|c|c}
        Context & Scor. & MAP@10 & stdev & best \%   \\ \hline
         {\bf Lin} & {\bf cent} & \bf{.78} & \bf{.22}  & \\ 
         List & csum & .59 & .30 & 20  \\
         Dep & cent & .53 & .31 & 15 \\
         SP & csum & .48 & .32 & 10  \\
         UP & csum & .47 & .36 & 10 \\
    \end{tabular}
    \caption{Comparison of the different context types. For each context type, we report the scoring method with higher MAP@10 on dev set, MAP@10 with 5 seed terms, its standard deviation among the different test term lists, the percentage of the test term lists where the context type achieves best performance.
    }
    \label{tab:context}
\end{table}

\subsection{Context Combination}
We provide in Table~\ref{tab:map} MAP@$n$ for the centroid scoring of the linear context and for the MLP classification with 5 seed terms. 
For comparison, we report in `Concat.' row the performance for the MLP binary classification on the concatenation of the multi-context term embeddings.
In addition, we report {\it oracle} performance assuming we have an oracle that chooses, for each term list, the best context type with the best scoring method. 
Oracle performance shows that the context types are indeed complementary. 
The MLP classifier which combines all the context types, yields additional improvement in the MAP@$n$ compared to the baseline linear context. 
Moreover, we observed that the improvement of the MLP combination over the linear context is preserved with 2 and 10 seed terms. Yet, looking at the oracle, the MLP combination still does not optimally integrate all the information captured by the term embeddings. 

\begin{table}
    \centering
    \begin{tabular}{l|c|c|c}
         Method & MAP@10 & MAP@20 & MAP@50  \\ \hline
         Linear & .78 &	.71 &	.59 \\
         Concat. & .68 & .65 & .56 \\
         {\bf MLP} & {\bf .83} &	{\bf .74} &	{\bf .63} \\ \hline
         Oracle & .89 & .82 & .73 \\ 
    \end{tabular}
    \caption{MAP@$n$ performance evaluation of the linear context, concatenation, MLP binary classification and oracle, with 5 seed terms.}
    \label{tab:map}
\end{table}

\section{Conclusion}
We proposed a novel approach to combine different context embedding types and we showed that it achieved improved results for the corpus-based term set expansion use case. In addition, we publish a dataset and a companion corpus that enable comparability and replicability of work in this field.

For future work, we plan to run similar experiments using recently introduced contextual embeddings, (e.g., ELMo~\cite{peters2018deep}, BERT~\cite{devlin2018bert}, OpenAI GPT-2~\cite{radford2019language}), which are expected to implicitly capture more syntax than context-free embeddings used in the current paper. We plan also to investigate the contribution of multi-context term embeddings to other tasks in computational semantics.

\bibliography{references.bib}

\begin{thebibliography}{30}
\expandafter\ifx\csname natexlab\endcsname\relax\def\natexlab#1{#1}\fi

\bibitem[{Berant et~al.(2012)Berant, Dagan, and
  Goldberger}]{berant2012learning}
J.~Berant, I.~Dagan, and J.~Goldberger. 2012.
\newblock Learning entailment relations by global graph structure optimization.
\newblock \emph{Computational Linguistics}, 38:73--111.

\bibitem[{Choi et~al.(2015)Choi, Tetreault, and Stent}]{choi2015depends}
Jinho~D Choi, Joel Tetreault, and Amanda Stent. 2015.
\newblock It depends: Dependency parser comparison using a web-based evaluation
  tool.
\newblock In \emph{Proceedings of the 53rd Annual Meeting of the Association
  for Computational Linguistics and the 7th International Joint Conference on
  Natural Language Processing (Volume 1: Long Papers)}, volume~1, pages
  387--396.

\bibitem[{Davidov and Rappoport(2006)}]{davidov2006efficient}
Dmitry Davidov and Ari Rappoport. 2006.
\newblock Efficient unsupervised discovery of word categories using symmetric
  patterns and high frequency words.
\newblock In \emph{Proceedings of the 21st International Conference on
  Computational Linguistics and the 44th annual meeting of the Association for
  Computational Linguistics}, pages 297--304. Association for Computational
  Linguistics.

\bibitem[{Demartini et~al.(2009)Demartini, Iofciu, and
  De~Vries}]{demartini2009overview}
Gianluca Demartini, Tereza Iofciu, and Arjen~P De~Vries. 2009.
\newblock Overview of the inex 2009 entity ranking track.
\newblock In \emph{International Workshop of the Initiative for the Evaluation
  of XML Retrieval}, pages 254--264. Springer.

\bibitem[{Devlin et~al.(2018)Devlin, Chang, Lee, and
  Toutanova}]{devlin2018bert}
Jacob Devlin, Ming-Wei Chang, Kenton Lee, and Kristina Toutanova. 2018.
\newblock Bert: Pre-training of deep bidirectional transformers for language
  understanding.
\newblock \emph{arXiv preprint arXiv:1810.04805}.

\bibitem[{Gyllensten and Sahlgren(2018)}]{gyllensten2018distributional}
Amaru~Cuba Gyllensten and Magnus Sahlgren. 2018.
\newblock Distributional term set expansion.
\newblock \emph{arXiv preprint arXiv:1802.05014}.

\bibitem[{Harris(1954)}]{harris1954distributional}
Zellig~S Harris. 1954.
\newblock Distributional structure.
\newblock \emph{Word}, 10(2-3):146--162.

\bibitem[{He and Xin(2011)}]{he2011seisa}
Yeye He and Dong Xin. 2011.
\newblock Seisa: set expansion by iterative similarity aggregation.
\newblock In \emph{Proceedings of the 20th international conference on World
  wide web}, pages 427--436. ACM.

\bibitem[{Joulin et~al.(2016)Joulin, Grave, Bojanowski, Douze, J{\'e}gou, and
  Mikolov}]{joulin2016fasttext}
Armand Joulin, Edouard Grave, Piotr Bojanowski, Matthijs Douze, H{\'e}rve
  J{\'e}gou, and Tomas Mikolov. 2016.
\newblock Fasttext.zip: Compressing text classification models.
\newblock \emph{arXiv preprint arXiv:1612.03651}.

\bibitem[{Kageura and Umino(1996)}]{kageura1996methods}
Kyo Kageura and Bin Umino. 1996.
\newblock Methods of automatic term recognition: A review.
\newblock \emph{Terminology. International Journal of Theoretical and Applied
  Issues in Specialized Communication}, 3(2):259--289.

\bibitem[{Levy and Goldberg(2014)}]{levy2014dependency}
Omer Levy and Yoav Goldberg. 2014.
\newblock Dependency-based word embeddings.
\newblock In \emph{Proceedings of the 52nd Annual Meeting of the Association
  for Computational Linguistics (Volume 2: Short Papers)}, volume~2, pages
  302--308.

\bibitem[{MacAvaney and Zeldes(2018)}]{macavaney2018deeper}
Sean MacAvaney and Amir Zeldes. 2018.
\newblock A deeper look into dependency-based word embeddings.
\newblock \emph{arXiv preprint arXiv:1804.05972}.

\bibitem[{Mamou et~al.(2018{\natexlab{a}})Mamou, Pereg, Wasserblat, Dagan,
  Goldberg, Eirew, Green, Guskin, Izsak, and Korat}]{mamou2018term}
Jonathan Mamou, Oren Pereg, Moshe Wasserblat, Ido Dagan, Yoav Goldberg, Alon
  Eirew, Yael Green, Shira Guskin, Peter Izsak, and Daniel Korat.
  2018{\natexlab{a}}.
\newblock {Term Set Expansion based on Multi-Context Term Embeddings: an
  End-to-end Workflow}.
\newblock In \emph{Proceedings of the 27th International Conference on
  Computational Linguistics: System Demonstrations}.

\bibitem[{Mamou et~al.(2018{\natexlab{b}})Mamou, Pereg, Wasserblat, Eirew,
  Green, Guskin, Izsak, and Korat}]{mamou2018term2}
Jonathan Mamou, Oren Pereg, Moshe Wasserblat, Alon Eirew, Yael Green, Shira
  Guskin, Peter Izsak, and Daniel Korat. 2018{\natexlab{b}}.
\newblock {Term Set Expansion based NLP Architect by Intel AI Lab}.
\newblock In \emph{Proceedings of the Conference on Empirical Methods in
  Natural Language Processing: System Demonstrations}.

\bibitem[{Mikolov et~al.(2013)Mikolov, Sutskever, Chen, Corrado, and
  Dean}]{mikolov2013distributed}
Tomas Mikolov, Ilya Sutskever, Kai Chen, Greg~S Corrado, and Jeff Dean. 2013.
\newblock Distributed representations of words and phrases and their
  compositionality.
\newblock In \emph{Advances in neural information processing systems}, pages
  3111--3119.

\bibitem[{Moreno and Redondo(2016)}]{moreno2016text}
Antonio Moreno and Te{\'o}filo Redondo. 2016.
\newblock Text analytics: the convergence of big data and artificial
  intelligence.
\newblock \emph{IJIMAI}, 3(6):57--64.

\bibitem[{Pantel et~al.(2009)Pantel, Crestan, Borkovsky, Popescu, and
  Vyas}]{pantel2009web}
Patrick Pantel, Eric Crestan, Arkady Borkovsky, Ana-Maria Popescu, and Vishnu
  Vyas. 2009.
\newblock Web-scale distributional similarity and entity set expansion.
\newblock In \emph{Proceedings of the 2009 Conference on Empirical Methods in
  Natural Language Processing: Volume 2-Volume 2}, pages 938--947. Association
  for Computational Linguistics.

\bibitem[{Pennington et~al.(2014)Pennington, Socher, and
  Manning}]{pennington2014glove}
Jeffrey Pennington, Richard Socher, and Christopher Manning. 2014.
\newblock Glove: Global vectors for word representation.
\newblock In \emph{Proceedings of the 2014 conference on empirical methods in
  natural language processing (EMNLP)}, pages 1532--1543.

\bibitem[{Peters et~al.(2018)Peters, Neumann, Iyyer, Gardner, Clark, Lee, and
  Zettlemoyer}]{peters2018deep}
Matthew~E Peters, Mark Neumann, Mohit Iyyer, Matt Gardner, Christopher Clark,
  Kenton Lee, and Luke Zettlemoyer. 2018.
\newblock Deep contextualized word representations.
\newblock \emph{arXiv preprint arXiv:1802.05365}.

\bibitem[{Radford et~al.(2019)Radford, Wu, Child, Luan, Amodei, and
  Sutskever}]{radford2019language}
Alec Radford, Jeff Wu, Rewon Child, David Luan, Dario Amodei, and Ilya
  Sutskever. 2019.
\newblock Language models are unsupervised multitask learners.

\bibitem[{Roark and Charniak(1998)}]{roark1998noun}
Brian Roark and Eugene Charniak. 1998.
\newblock Noun-phrase co-occurrence statistics for semiautomatic semantic
  lexicon construction.
\newblock In \emph{Proceedings of the 36th Annual Meeting of the Association
  for Computational Linguistics and 17th International Conference on
  Computational Linguistics-Volume 2}, pages 1110--1116. Association for
  Computational Linguistics.

\bibitem[{Rong et~al.(2016)Rong, Chen, Mei, and Adar}]{rong2016egoset}
Xin Rong, Zhe Chen, Qiaozhu Mei, and Eytan Adar. 2016.
\newblock Egoset: Exploiting word ego-networks and user-generated ontology for
  multifaceted set expansion.
\newblock In \emph{Proceedings of the Ninth ACM International Conference on Web
  Search and Data Mining}, pages 645--654. ACM.

\bibitem[{Sarmento et~al.(2007)Sarmento, Jijkuon, de~Rijke, and
  Oliveira}]{sarmento2007more}
Luis Sarmento, Valentin Jijkuon, Maarten de~Rijke, and Eugenio Oliveira. 2007.
\newblock More like these: growing entity classes from seeds.
\newblock In \emph{Proceedings of the sixteenth ACM conference on Conference on
  information and knowledge management}, pages 959--962. ACM.

\bibitem[{Schwartz et~al.(2015)Schwartz, Reichart, and
  Rappoport}]{schwartz2015symmetric}
Roy Schwartz, Roi Reichart, and Ari Rappoport. 2015.
\newblock Symmetric pattern based word embeddings for improved word similarity
  prediction.
\newblock In \emph{Proceedings of the Nineteenth Conference on Computational
  Natural Language Learning}, pages 258--267.

\bibitem[{Shaw et~al.(1994)Shaw, Fox, Shaw, and Fox}]{Fox94combinationof}
Joseph~A. Shaw, Edward~A. Fox, Joseph~A. Shaw, and Edward~A. Fox. 1994.
\newblock Combination of multiple searches.
\newblock In \emph{The Second Text REtrieval Conference (TREC-2}, pages
  243--252.

\bibitem[{Shen et~al.(2017)Shen, Wu, Lei, Shang, Ren, and
  Han}]{shen2017setexpan}
Jiaming Shen, Zeqiu Wu, Dongming Lei, Jingbo Shang, Xiang Ren, and Jiawei Han.
  2017.
\newblock Setexpan: Corpus-based set expansion via context feature selection
  and rank ensemble.
\newblock In \emph{Joint European Conference on Machine Learning and Knowledge
  Discovery in Databases}, pages 288--304. Springer.

\bibitem[{Shi et~al.(2010)Shi, Zhang, Yuan, and Wen}]{shi2010corpus}
Shuming Shi, Huibin Zhang, Xiaojie Yuan, and Ji-Rong Wen. 2010.
\newblock Corpus-based semantic class mining: distributional vs. pattern-based
  approaches.
\newblock In \emph{Proceedings of the 23rd International Conference on
  Computational Linguistics}, pages 993--1001. Association for Computational
  Linguistics.

\bibitem[{Wu et~al.(2006)Wu, Crestani, and Bi}]{wu2006evaluating}
Shengli Wu, Fabio Crestani, and Yaxin Bi. 2006.
\newblock Evaluating score normalization methods in data fusion.
\newblock In \emph{Asia Information Retrieval Symposium}, pages 642--648.
  Springer.

\bibitem[{Zaheer et~al.(2017)Zaheer, Kottur, Ravanbakhsh, Poczos,
  Salakhutdinov, and Smola}]{zaheer2017deep}
Manzil Zaheer, Satwik Kottur, Siamak Ravanbakhsh, Barnabas Poczos, Ruslan~R
  Salakhutdinov, and Alexander~J Smola. 2017.
\newblock Deep sets.
\newblock In \emph{Advances in Neural Information Processing Systems}, pages
  3394--3404.

\bibitem[{Zhao et~al.(2018)Zhao, Feng, Luo, and
  Tian}]{Zhao:2018:ESE:3234944.3234966}
He~Zhao, Chong Feng, Zhunchen Luo, and Changhai Tian. 2018.
\newblock Entity set expansion from twitter.
\newblock In \emph{Proceedings of the 2018 ACM SIGIR International Conference
  on Theory of Information Retrieval}, ICTIR '18, pages 155--162, New York, NY,
  USA. ACM.

\end{thebibliography}
\bibliographystyle{acl_natbib_nourl}

\end{document}